\documentclass[10pt]{article}
\usepackage[margin=0.85in]{geometry}
\usepackage{amsmath,amssymb}
\usepackage{booktabs}
\usepackage{graphicx}
\usepackage{float}
\graphicspath{{figures/}{../figures/}}
\usepackage{xcolor}
\usepackage[hidelinks]{hyperref}
\usepackage{natbib}
\usepackage{caption}
\newcommand{\sigsq}[1]{\sigma^2_{#1}}
\newcommand{\erho}{E\rho^2}

\title{\textbf{Ask Which, Not How Good:\\ Sizing Benchmarks Scored by an LLM}}
\author{Atul Anand}
\date{}

\begin{document}
\maketitle

\begin{abstract}
Benchmarks scored by an LLM judge are used to adjudicate differences of a tenth of a
point, but the resolution of those benchmarks has never been measured. Existing
sample-complexity work covers \emph{accuracy} benchmarks and leaves the judged case open.
We close it. Treating the \emph{system} as the object of measurement, we decompose
350{,}000 judgments into system, item, judge and interaction components, holding a
9-judge panel, one pointwise 0--5 rubric and one scale fixed while varying the
items, drawn from MT-Bench, Arena-Hard, AlpacaEval~2 and a discrimination-screened set.
System panels are close but not identical across item sets, in two ways we record.

Three results follow. First, a structural one: under a single judge, generalizability
asymptotes to $\sigsq{s}/(\sigsq{s}+\sigsq{sj})$ \emph{regardless of item count}, because
the system-by-judge term carries no $n_i$. Items saturate; judges do not, and the item
cost of a target \emph{diverges} as the target nears the ceiling: on MT-Bench under its own
protocol, \textbf{two judges at thirty items match 77 items at one judge} at
$\erho=0.75$, while at $0.80$---within $0.006$ of that benchmark's ceiling---no item count
suffices at all.
Second, \emph{protocol design moves the ceiling itself}: on identical prompts, MT-Bench's
native template, 1--10 scale and reference answers double the share of variance
attributable to the system (4.8\% to 8.4\% on matched material) and cut the judges
required from 18 to 3, at no cost per call. Third, minimum detectable differences are 0.41--1.24 points on a 0--5
scale at native item counts, while the median of 60 improvements recovered from published
papers is \textbf{0.28 points}; on the one benchmark recurring often enough for an exactly
matched comparison, \textbf{all 17 recovered MT-Bench improvements fall below MT-Bench's
own floor}.

The \emph{existence} of the ceiling is invariant; its \emph{height} is not. We take the six
current-generation judges as the primary universe, since the other three were included to
probe the judge facet rather than because anyone would hire them, and report the nine-judge
panel as sensitivity. That concession costs the alarm and not the argument: under the
primary universe three of four item sets clear $\erho=0.8$ and MT-Bench's ceiling rises
from $0.600$ to $0.798$, yet a single judge still cannot reach $0.8$ on it at any item
count, and still needs five judges to do so.

\emph{Protocol matters more than panel size, and we ran the protocol most of the field
uses.} Under a pairwise preference---both presentation orders, position as an estimable
facet---$\sigsq{sj}$ falls two orders of magnitude below $\sigsq{s}$ and the ceiling rises
to $\mathbf{0.986}$ (bootstrap $[0.934,1.000]$ on 11 systems; $0.954$--$0.974$ restricted to
current models), so one judge suffices.
The ceiling result is therefore a property of \emph{pointwise rubric scoring}, not of LLM
judging as such. Pairwise buys its own problem instead: a system presented first wins
\textbf{8.6 percentage points} more often than the same system presented second, which is
$1.23\times$ the median improvement claimed in the 53 published win-rate comparisons we
recovered, and $70\%$ of those fall below the pairwise floor at native item counts.

An audit of 628 arXiv papers---double-coded by two independent models and validated
against blind human coding ($\kappa=0.73$)---finds that fewer than one paper in four
states whether its evaluation was run more than once (16.7\% human, 20.0\% automated,
$\kappa=0.89$), and that only 46--67\% report uncertainty of any kind. The crossed
judgment dataset, a D-study calculator and the audit codebook are available from the
author on request.
\end{abstract}

\section{Introduction}

A large fraction of reported progress in language modelling is adjudicated by an LLM
judge. A method is declared better because it scores 7.4 rather than 7.2 on MT-Bench,
or wins 34.5\% rather than 31.8\% on AlpacaEval. These are measurements, and like all
measurements they have a resolution, but that resolution is almost never reported and,
for judged benchmarks, has not been characterised.

For \emph{accuracy} benchmarks the question is settled in outline.
\citet{miller2024errorbars} sets out the statistical case for reporting uncertainty on
benchmark results and supplies the item-sampling machinery; a subsequent industry analysis
\citep{tmls2026samplecomplexity} applies it to compute minimum detectable effects on MMLU,
HumanEval and GSM8K and reports that many published gaps fall inside their own benchmark's
noise. Both treat evaluations whose only noise sources are item sampling and generation,
and both identify the judged case---where a judge facet with its own behaviour is
added---as open. This paper is that formula, and the measurement that
goes with it. The framing follows \citet{schaeffer2023mirage} in treating a believed
empirical claim as a measurement artefact.

The judge facet is not merely one more variance source. Writing the measurement model
out (\S\ref{sec:model}) shows that the system-by-judge interaction enters the error
variance as $\sigsq{sj}/n_j$, a term containing no $n_i$. Items therefore cannot
reduce it. This has an immediate and, to our knowledge, unremarked consequence: a
single-judge protocol has a reliability ceiling that no amount of data collection can
raise. Whether that ceiling sits above or below a usable threshold is an empirical
question, and we answer it for the benchmarks the field actually publishes on.

\paragraph{Contributions.}
\begin{enumerate}\itemsep2pt
\item \textbf{A sample-complexity result for judged benchmarks.} Under one judge,
$\erho \to \sigsq{s}/(\sigsq{s}+\sigsq{sj})$ as $n_i\to\infty$, so items saturate and
judges do not. The exchange rate steepens as the target nears the ceiling: two judges at
thirty items match 77 items at one judge at $\erho=0.75$, and no item count suffices at
$0.80$. The saturation is universe-invariant; whether a given ceiling clears a usable
threshold is not; we take the six current-generation judges as the primary universe and
report the wider panel as sensitivity (\S\ref{sec:ceiling}, \S\ref{sec:primary-universe}).
\item \textbf{The ceiling is a property of pointwise rubric scoring, not of LLM judging.}
Run as a pairwise preference in both presentation orders, $\sigsq{sj}$ falls two orders of
magnitude below $\sigsq{s}$, the ceiling rises to 0.986 ($[0.934,1.000]$, 11 systems), and
one judge suffices, but a
system presented first wins 8.6 percentage points more often than the same system
presented second, a bias $1.23\times$ the median published win-rate claim. With that floor
measured, $70\%$ of the 53 win-rate comparisons we recovered fall below it
(\S\ref{sec:pairwise}).
\item \textbf{Protocol design moves the ceiling.} On identical prompts, MT-Bench's native
protocol raises $\sigsq{s}$ share from 4.8\% to 8.4\% against a generic holistic rubric and
cuts the judges needed for $\erho=0.8$ from 18 to 3, on matched material. Rubric and scale are a cheaper lever than either
items or judges, and are almost never treated as a design variable
(\S\ref{sec:protocol}).
\item \textbf{Measured noise floors} of 0.41--1.24 points on a 0--5 scale at native item
counts, with a common judge panel, rubric and scale---system panels differ only as
\S\ref{sec:limits} records---so cross-set differences are attributable to the items; and a measured distribution of reported improvements whose median
(0.28 points) sits below every floor we measured---all 17 recovered MT-Bench improvements
fall below MT-Bench's own (\S\ref{sec:floors}, \S\ref{sec:deltas}).
\item \textbf{Reliability is a property of the (benchmark, judge-universe) pair.}
MT-Bench's ceiling moves between 0.600 and 0.798 depending on which judges are admitted;
no published work states its judge universe (\S\ref{sec:universe}).
\item \textbf{The limit is a property of the design, not of LLM judges.} On a matched
design LLM judges are \emph{more} self-consistent than human experts and far more so than
crowd workers (\S\ref{sec:human}); and discriminating power collapses when the systems
compared are close, system variance falling to 1.9--4.2\% on three of four item sets once
restricted to ten current frontier models (\S\ref{sec:frontier}).
\item \textbf{Item-discrimination screening is the one lever that raises $\sigsq{s}$.}
Screening candidate items by between-system variance moved $\sigsq{s}$ by roughly two
orders of magnitude in a pilot, and yields a system-variance share comparable to the best
public set (16.6\% against Arena-Hard's 19.2\%, on items chosen for that property rather
than found to have it). Since items saturate, this is the only axis that raises the
ceiling rather than approaching it (\S\ref{sec:setup}).
\item \textbf{Judge provenance is a further facet.} Replicating the design with five
open-weight judges on identical material, judges disagree $2.17\times$ more across model
families than within them once scale usage is standardised away, so panel size overstates
panel independence, though mixing families does not reduce the judges required
(\S\ref{sec:provenance}).
\item \textbf{An audit of 628 papers}, double-coded by two independent models, showing
that the disclosure needed to assess any of this is largely absent, and quantifying how
reliably each disclosure can be coded at all (\S\ref{sec:audit}).
\end{enumerate}

\section{Related work}\label{sec:related}

\paragraph{Sizing and uncertainty for benchmarks.} \citet{miller2024errorbars} makes the
case for reporting uncertainty on evaluation results and supplies the item-sampling
machinery; \citet{tmls2026samplecomplexity} applies it to compute minimum detectable
effects on accuracy benchmarks and finds many published gaps inside their own noise. Both
treat evaluations whose only stochastic facets are item sampling and generation, and both
identify the judged case as open. Within NLP, \citet{card2020power} showed that common
experimental designs are underpowered for the effects they claim, \citet{dror2018hitchhiker}
set out significance-testing practice for the field, and \citet{deutsch2021statistical}
gave confidence intervals for summarisation metrics that account for both systems and
items. Our contribution is to add the judge facet, which is what changes the asymptotics:
item sampling can be beaten by collecting more items and the judge facet cannot.
\citet{schaeffer2023mirage} is the closest in spirit, treating a believed empirical claim
as an artifact of the measurement rather than a property of the model.

\paragraph{Generalizability theory.} The variance-decomposition machinery is standard in
educational measurement \citep{cronbach1972dependability,brennan2001generalizability,
shavelson1991generalizability}, where the object of measurement is normally a person and
raters are the nuisance facet. Recent applications to LLM evaluation
\citep{hiddenmeasurement2026,autoscoring2025} keep the item as the object and therefore
answer an annotation-quality question. Taking the \emph{system} as the object is what makes
$\sigsq{sj}/n_j$ the binding term, since it is the interaction between the thing being
measured and the instrument measuring it.

\paragraph{Judge reliability and bias.} A substantial literature documents that LLM judges
are noisy and biased evaluators: \citet{coinflipjudge2026} and
\citet{reliabilitywithoutvalidity2026} characterise agreement, consistency and bias across
judge models; \citet{yang2026judgechanges} shows that conclusions move when the judge
changes; \citet{judgmentnoise2025} attributes much of the problem to design failures in
judge benchmarks; and \citet{irtjudge2026} applies item response theory to diagnose judge
quality. This work establishes \emph{that} judges disagree. What it does not derive is the
consequence for sample size, which is the gap this paper fills: disagreement between judges
enters the error variance in a term that no amount of item collection reduces.

\paragraph{Benchmarks and leaderboards.} We measure items drawn from MT-Bench
\citep{zheng2023mtbench}, Arena-Hard \citep{li2024arenahard} and AlpacaEval~2
\citep{dubois2024alpacaeval}. Work on leaderboard stability under resampling and on the
statistical properties of Elo-style aggregation is complementary to ours: it asks how a
ranking moves under perturbation, where we ask how finely the underlying scores can be
resolved at all. \citet{bowman2021will} argues that benchmark progress claims outrun the
evidence supporting them, which is the concern our instruments are intended to make
checkable.

\section{The measurement model}\label{sec:model}

We treat the \emph{system} as the object of measurement, in the generalizability-theory
sense of \citet{cronbach1972dependability} and \citet{brennan2001generalizability}. This
differs from prior applications of G-theory to LLM evaluation
\citep{hiddenmeasurement2026,autoscoring2025}, which take the item as the object and
therefore answer an annotation-quality question rather than a benchmark-validity one.
Related measurement work studies judges through item response theory
\citep{irtjudge2026}, replicate-aggregation curves \citep{coinflipjudge2026}, judge
substitution \citep{yang2026judgechanges}, and design failures in judge benchmarks
\citep{judgmentnoise2025,reliabilitywithoutvalidity2026}; none derives or measures the
item-count ceiling below. A score is modelled as
\begin{equation}
X_{sijr}=\mu+\nu_s+\nu_i+\nu_j+\nu_{si}+\nu_{sj}+\nu_{ij}+\nu_{sij}+\nu_{r(sij)} .
\end{equation}
With $r=1$, $\sigsq{sij}$ and the residual are not separately identified: \emph{a
single-run design cannot compute its own noise floor}. We therefore use $r=3$.

Relative error variance, generalizability, and the minimum detectable difference between
two systems measured on the same items and judges are
\begin{align}
\sigsq{\delta} &= \frac{\sigsq{si}}{n_i}+\frac{\sigsq{sj}}{n_j}
  +\frac{\sigsq{sij}}{n_i n_j}+\frac{\sigsq{e}}{n_i n_j n_r},\qquad
\erho = \frac{\sigsq{s}}{\sigsq{s}+\sigsq{\delta}},\\
\mathrm{MDD}_{95} &\approx 1.96\sqrt{2\,\sigsq{\delta}} .
\end{align}

\paragraph{The ceiling.}
Setting $n_j=1$ and letting $n_i\to\infty$, every term of $\sigsq{\delta}$ vanishes
except $\sigsq{sj}$, giving
\begin{equation}\label{eq:ceiling}
\lim_{n_i\to\infty}\erho\big|_{n_j=1}=\frac{\sigsq{s}}{\sigsq{s}+\sigsq{sj}},
\qquad
\lim_{n_i\to\infty}\mathrm{MDD}_{95}\big|_{n_j=1}=1.96\sqrt{2\,\sigsq{sj}} .
\end{equation}
Items buy nothing against $\sigsq{sj}$; only judges do. Equation~\eqref{eq:ceiling}
follows directly from standard D-study algebra \citep{brennan2001generalizability}; our
claim is not the derivation but that its consequence has never been measured for judged
benchmarks, nor drawn. ``Run more items'' is the field's reflexive response to
evaluation noise, and against this component it is futile.

\paragraph{Why MDD rather than $\erho$.}
$\sigsq{\delta}$ is built from components estimated on tens of thousands of degrees of
freedom, whereas $\sigsq{s}$ has only $n_s-1$. MDD contains no $\sigsq{s}$; $\erho$
divides by it. In our data the bootstrap interval on MDD is roughly $\pm7\%$ of its
point estimate while that on $\sigsq{s}$ exceeds $100\%$. We therefore report MDD as the
primary quantity and $\erho$ as an interval.

\section{Experimental setup}\label{sec:setup}

\paragraph{Design.} 42 systems (14 models $\times$ 3 prompt configurations, spanning
2024-era to current frontier), $n_i$ items per benchmark, 9 judges, 3 replicates,
pointwise scoring on a 0--5 scale. The same judges, rubric and scale are used across all four item sets. The system panels are
close but not identical, and \S\ref{sec:limits} records the two ways they differ, so
cross-set differences are attributable to the items up to that caveat rather than
unconditionally.

Two system counts appear in the tables and the difference is deliberate. The three public
item sets are scored on the 42 systems above. The screened set additionally carries a
\emph{null arm}: for twelve of the systems we drew two further independent generation
samples under an identical configuration, giving pairs whose true difference is exactly
zero, which is what makes the Type~I error check in \S\ref{sec:confounds} possible. Those
replicas are systems for the purposes of the decomposition, so the screened set has
$36+23=59$ after one replica is dropped for incomplete coverage. Its base panel is 36
rather than 42 because \texttt{claude-opus-5} and \texttt{claude-sonnet-5} were not
available as systems when the screened set was collected, in any of the three prompt
configurations; both remain in the judge panel. Refitting the screened row on those 36
base systems alone changes it little (MDD 1.25 against 1.24, ceiling 0.735 against 0.746),
so the cross-set comparison in Table~\ref{tab:main} does not turn on the null arm. The near-frontier subset
(Appendix~\ref{sec:frontier}) is the ten current frontier models at three configurations,
plus their null-arm replicas where present, giving 38 on the screened set and 30
elsewhere.

\paragraph{Judges.} A crossed 2 families $\times$ 3 tiers panel of current models, plus
three further judges: a deliberately weak judge, a legacy judge representing what
published work typically used, and a \emph{route replica}, the same weights served
through a different provider route.

All three are \emph{included} in the nine-judge universe that Table~\ref{tab:main} reports.
We report that universe as the headline because published work states no universe at all
and a reader assembling ``some LLM judges'' could plausibly land on it. The choice is not
neutral---a universe containing a deliberately weak judge carries more system-by-judge
disagreement, hence a lower ceiling and a larger judge requirement, which is the direction
that flatters this paper's thesis---and it turns out to be load-bearing rather than
cosmetic. We therefore take the \emph{six current-generation judges} as the primary
universe for interpretation, since it is what a practitioner assembling a panel today would
build, and report the nine-judge fit in the tables because it is what a reader
reconstructing our full design would compute. Under the primary universe three of the four
item sets clear $\erho=0.8$; under the nine-judge universe none does. The structural result
is identical either way. \S\ref{sec:ceiling} quantifies the gap and \S\ref{sec:universe}
gives every headline quantity under four universes.

\paragraph{Benchmarks.} MT-Bench \citep{zheng2023mtbench}, Arena-Hard
\citep{li2024arenahard} and AlpacaEval~2 \citep{dubois2024alpacaeval} items are random
samples under a fixed seed. A fourth set is constructed by \emph{item discrimination screening}:
candidates are scored by a held-out judge and retained by between-system variance, which
is each item's contribution to $\sigsq{s}$. In a 15-system pilot this moved $\sigsq{s}$
from $0.0027$ to $0.6958$, roughly two orders of magnitude; the pilot is small and its two
arms differ in item count as well as in screening, so we take the direction and the order
of magnitude rather than a point estimate. What the screened set buys in the main study,
and what it costs, is reported with the other item sets in \S\ref{sec:floors}.

The motivation is that classical item analysis has screened items for discrimination for a
century while benchmark construction generally does not, selecting instead for difficulty,
diversity or realism. A benchmark can therefore carry hundreds of items and almost no
system signal, and because items saturate (\S\ref{sec:ceiling}), adding more cannot
repair it. Screening is the only lever we tested that raises $\sigsq{s}$ itself, which is
the numerator every other quantity in this paper depends on.

\paragraph{Infrastructure.} Three failure modes materially affected measurement and are
reported because they generalise. (i) The inference gateway performed \emph{response
caching}: 99.8\% of replicate cells initially returned a byte-identical cached response,
collapsing the replicate facet and making $\sigsq{e}$ read as zero. Any judge-reliability
study run through a caching gateway will conclude judges are perfectly self-consistent.
(ii) Reasoning-model judges silently returned empty completions when the token budget
was sized for a score, removing a judge from the panel without removing it from the
reported design. (iii) Prompt caching did not function, so replicate cost is not
discounted. All calls log model version, timestamp, prompt hash and response
identifier; the response identifier is what made (i) detectable.

\section{Results}

\subsection{Noise floors}\label{sec:floors}

Table~\ref{tab:main} reports the decomposition. Minimum detectable differences at a common
thirty items, one judge and one run range from 0.58 to 1.24 points on a 0--5 scale, and
system variance is 4.1--4.5\% of total score variance on MT-Bench and AlpacaEval~2: the
overwhelming majority of what a judge score varies with is not the system.

Table~\ref{tab:main} is fitted on all nine judges, which is the \emph{sensitivity} universe,
not the primary one: three of the nine were included to probe the judge facet rather than
because anyone would hire them. Section~\ref{sec:primary-universe} sets out the primary universe of
six current-generation judges and reports every ceiling under both, and readers who want
the headline numbers for a panel they would actually assemble should read
Table~\ref{tab:universe-primary} first. We present the nine-judge fit here because the rest
of this section varies the item facet against a fixed judge set, and the wider set makes
that comparison harder rather than easier on us.

\begin{table}[t]\centering\small
\caption{Detectability and the single-judge ceiling. MDD and $\erho$ at a common $n_i=30$, one judge, one run, so benchmarks are compared at an equal item budget. Intervals are parametric bootstrap (600--800 resamples); $P(\ge\!0.8)$ is the bootstrap probability that the ceiling reaches the conventional threshold. The final column is the number of judges needed to reach $\erho=0.8$ at 30 items and 3 replicates.}
\label{tab:main}
\begin{tabular}{lrrrcrrcrr}
\toprule
Benchmark & $n_s$ & $n_i$ & MDD & 95\% CI & $\erho$ & ceiling & 95\% CI & $P(\ge\!0.8)$ & judges \\
\midrule
Screened (ours) & 59 & 30 & 1.24 & [1.17,\,1.31] & 0.695 & \textbf{0.746} & [0.65,\,0.81] & 6.2\% & 2 \\
MT-Bench & 42 & 42 & 0.63 & [0.60,\,0.66] & 0.463 & \textbf{0.600} & [0.46,\,0.72] & 0.0\% & 25 \\
Arena-Hard & 42 & 17 & 0.71 & [0.65,\,0.76] & 0.737 & \textbf{0.792} & [0.69,\,0.86] & 41.0\% & 2 \\
AlpacaEval\,2 & 42 & 32 & 0.58 & [0.54,\,0.61] & 0.542 & \textbf{0.706} & [0.55,\,0.80] & 2.0\% & 7 \\
\bottomrule
\end{tabular}
\end{table}

\subsection{Items cannot lift the ceiling}\label{sec:ceiling}

\begin{figure}[t]\centering
\includegraphics[width=.66\linewidth]{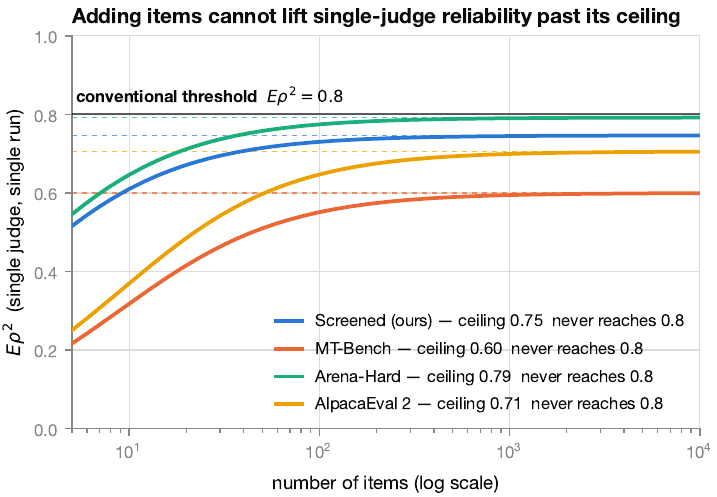}
\caption{Generalizability against item count under a single judge. Dashed lines are the
asymptotes of Eq.~\eqref{eq:ceiling}; the solid horizontal line is the conventional
$\erho=0.8$ threshold. All four flatten below it \emph{under the nine-judge universe};
under the six current-generation judges three of the four asymptote above it
(\S\ref{sec:primary-universe}).}
\label{fig:ceiling}
\end{figure}

Figure~\ref{fig:ceiling} plots $\erho$ against item count under a single judge. Every
curve flattens below the 0.8 threshold. The asymptotes are 0.600 (MT-Bench), 0.706
(AlpacaEval~2), 0.746 (our screened items) and 0.792 (Arena-Hard). Parametric bootstrap
intervals (800 resamples) put the posterior probability that the ceiling reaches 0.8 at
0.0\% for MT-Bench, 6.2\% for our screened set and 2.0\% for AlpacaEval~2. \textbf{On
these three, a single-judge protocol does not reach conventional reliability at any item
count under this judge universe.} Arena-Hard is the honest exception: at 0.792 with a
95\% interval of $[0.685,\,0.858]$ it has a 41\% probability of clearing the threshold,
and we do not claim it fails. The corresponding floors on MDD at infinite items are
0.40--1.09 points.

\paragraph{The exchange rate between the two axes.} Reaching $\erho=0.8$ at 30 items
requires 25 judges for MT-Bench, 7 for AlpacaEval~2, and 2 each for our screened set and
Arena-Hard (Table~\ref{tab:main}). Adding a \emph{second} judge at 30 items exceeds what a
single judge achieves at \emph{any} item count on three of the four benchmarks, because the
target sits above the single-judge ceiling and no amount of item collection reaches it. On
the fourth, AlpacaEval~2, a single judge needs 153 items to match two judges at 30, a
target far enough below that benchmark's ceiling ($0.666$ against $0.706$) for the count to
be stable, unlike the near-ceiling case in \S\ref{sec:protocol}. A second judge is not a
marginal improvement over more items; on most benchmarks it is the only thing that works.

These asymptotes are computed over the nine-judge universe; over the six current-generation
judges the same benchmarks give $0.929$, $0.918$, $0.829$ and $0.798$, so three of four
clear the threshold and only MT-Bench does not (\S\ref{sec:primary-universe}). The
saturation itself is universe-invariant---no ceiling is reachable by adding items, for any
universe, because $\sigsq{sj}/n_j$ carries no $n_i$---but whether a given ceiling is high
enough to be usable is a joint property of the benchmark and the judges admitted. We regard
that sensitivity as a finding rather than a caveat: a reliability number reported without a
stated judge universe is not conservative or liberal, it is undetermined.

\subsection{Protocol design moves the ceiling}\label{sec:protocol}

Everything above holds the judging protocol fixed---our generic holistic 0--5 rubric---so
that differences are attributable to items. That isolates the item effect but leaves a
question a reader will reasonably ask: are these properties of the benchmark, or of our
rubric? We answer it for MT-Bench, the one pointwise benchmark whose protocol we can
reproduce exactly, by re-running the same prompts under FastChat's own
\texttt{single-v1} template, its 1--10 scale, and the GPT-4 reference answers it supplies
for the math, reasoning and coding categories.

The two arms did not survive collection equally, and the difference runs against us. They
drew from different item pools---43 candidates for the generic arm, 50 for the native
one---and under the native protocol's longer prompts the \texttt{oai-cheap} judge returned
empty completions often enough to be dropped entirely, taking 12 of its 50 items with it.
As collected that leaves $38\times38\times8$ against $42\times42\times9$. The arm favouring
our conclusion is therefore also the arm with the smaller panel, which is exactly the shape
of confound a reader should distrust. We remove it by refitting both arms on the 38
systems, 33 items and 8 judges common to the two (Table~\ref{tab:protocol}, lower block)
and report that as the result.

Matched, MT-Bench under its native protocol recovers \textbf{8.4\% of variance as system
difference against 4.8\% under our rubric}---close to double---and its single-judge ceiling
rises from 0.623 to 0.798. The judges required for $\erho=0.8$ at thirty items fall from 18
to 3. Matching moves the generic arm slightly in its own favour (4.5\% to 4.8\%, ceiling
0.600 to 0.623) and the native arm slightly against it (9.1\% to 8.4\%, 0.804 to 0.798), so
the lever is not an artefact of the dropped judge.

\paragraph{The item exchange rate diverges near the ceiling.} That last shift is small but
it crosses a threshold, and the crossing is instructive. As collected, the native arm's
ceiling is 0.804 and a single judge reaches $\erho=0.8$ at 860 items; matched, the ceiling
is 0.798 and \emph{no} item count reaches 0.8. Both are true of the same protocol. The
reason is that $\erho$ approaches its asymptote hyperbolically, so the items required to
hit a target diverge as the target approaches the ceiling, and 0.8 sits within 0.006 of
this one. Away from the asymptote the exchange rate is stable and far less dramatic: at
$\erho=0.75$ the matched native arm needs 2 judges at thirty items or 77 items at one
judge, and at 0.78, 3 judges or 209 items. We therefore report the exchange rate as a
property that \emph{steepens} rather than as a single ratio, and treat any specific item
count quoted near a ceiling---including our own 860---as an artefact of proximity to the
asymptote rather than a stable quantity.

Three consequences. First, the cross-benchmark comparison in \S\ref{sec:floors} is
\emph{conservative}: holding a generic rubric fixed understates what each benchmark
achieves under its own design, so the true floors are likely lower than
Table~\ref{tab:main} reports. Second, rubric and scale are a design lever that costs
nothing per call---reference answers and a wider response scale bought more discriminating
power here than tripling the judge panel would have---and protocol design is almost never
reported as a choice, let alone justified. Third, it revises the strongest form of our
headline. Under a generic rubric no item count reaches $\erho=0.8$ on any item set studied;
under the native protocol the answer depends on which fit you read, which is precisely why
we state the surviving claim in its weaker form: a protocol change buys more than any
feasible item budget, and the budget required explodes near whichever ceiling the protocol
produces.

\begin{table}[t]\centering\small\setlength{\tabcolsep}{4pt}
\caption{Protocol design on identical MT-Bench prompts: our generic holistic 0--5 rubric versus MT-Bench's own FastChat template, 1--10 scale, and GPT-4 reference answers for math, reasoning and coding. The two arms drew from different item pools (43 and 50 candidates) and did not survive collection equally---under the native protocol's longer prompts the \texttt{oai-cheap} judge returned empty completions often enough to be dropped, along with 12 of its 50 items---so the upper block reports each arm as collected and the lower block refits both on the 38 systems, 33 items and 8 judges common to the two, which is the contrast the text relies on. MDD is on a common 0--5 basis. Judges is the count needed for $\erho=0.8$ at 30 items and 3 replicates; items the count at one judge and a single run, which diverges as the target approaches the ceiling.}
\label{tab:protocol}
\begin{tabular}{lccccccc}
\toprule
Protocol & design & $\sigsq{s}$ & $\erho$@30 & ceiling & MDD$_{0\text{-}5}$ & judges & items \\
\midrule
\multicolumn{8}{l}{\emph{as collected}}\\
Generic 0--5 rubric & 42$\times$42$\times$9 & 4.5\% & 0.463 & 0.600 & 0.63 & 25 & never \\
MT-Bench native & 38$\times$38$\times$8 & 9.1\% & 0.700 & 0.804 & 0.56 & 3 & 860 \\
\addlinespace
\multicolumn{8}{l}{\emph{matched: common 38 systems, 33 items, 8 judges}}\\
Generic 0--5 rubric & 38$\times$33$\times$8 & 4.8\% & 0.483 & 0.623 & 0.61 & 18 & never \\
MT-Bench native & 38$\times$33$\times$8 & \textbf{8.4\%} & 0.685 & \textbf{0.798} & 0.56 & \textbf{3} & never \\
\bottomrule
\end{tabular}
\end{table}

\subsection{The pairwise protocol, and where the ceiling goes}\label{sec:pairwise}

Everything to this point scores one response at a time against a rubric. Most of the field
does not: Arena-Hard and AlpacaEval~2 are pairwise, Chatbot Arena is pairwise, and the 53
win-rate comparisons we recovered from published papers are pairwise. A result about
pointwise rubrics that is silent on pairwise judging describes a minority of practice. So
we ran it.

\paragraph{Design.} Each of 12 systems is compared against a fixed baseline
(\texttt{gpt-4o-mini}) on the same 24 items by the same 4 judges, twice, with the judge
returning a preference rather than a score. The whole grid is run in \emph{both}
presentation orders---system first and baseline first---so position is an estimable facet
rather than a nuisance averaged away. Responses are reused from the main harvest, so this
arm costs judging only. The crossed decomposition is unchanged and the ceiling derivation
carries over verbatim, because $\sigsq{sj}/n_j$ contains no $n_i$ whatever the judge
returns.

\paragraph{The ceiling largely disappears.} Order-balanced, the system term is $35.7\%$ of
total variance and the single-judge ceiling is $\mathbf{0.986}$; one judge reaches
$\erho=0.93$ at 24 items. This is not a saturation artefact---no system's win rate is
pinned at 0 or 1, the observed range being $0.10$ to $0.87$---but the 12 systems do span
gpt-3.5-turbo to gpt-5.4-mini, and Section~\ref{sec:frontier} shows that range inflates
$\sigsq{s}$. Dropping the 2023 models halves the system share, to $17.5$--$19.8\%$, and the
ceiling still comes back at $0.954$--$0.974$ with one judge reaching $\erho\approx0.86$.
The effect survives the restriction that kills it in the pointwise design.

This arm is small, and the precision follows. After reduction to a complete block it carries
11 systems against the main study's 42, so $\sigsq{s}$ rests on 10 degrees of freedom rather
than 41, and it is the numerator of the ceiling. The parametric bootstrap gives $[0.934,
1.000]$ for the balanced ceiling and $[0.824, 0.983]$ for the tighter frontier-only fit
(truncated at 1, which the estimator does not enforce). Those are wide intervals and we do
not want them read as precise. What carries the claim is that the \emph{lower} bounds,
$0.934$ and $0.824$, still sit above every pointwise ceiling we measured
($0.600$--$0.798$): the gap between protocols is larger than the uncertainty in the
pairwise estimate, even though that uncertainty is substantial.

We state the consequence plainly, because it cuts against our own headline. Under a binary
preference the judges barely disagree: $\sigsq{sj}$ is two orders of magnitude below
$\sigsq{s}$, where under our generic 0--5 rubric it is comparable to it. Asking ``which of
these two is better'' is a far better conditioned question than asking ``what score does
this deserve,'' and the ceiling result---no item count suffices---is a property of
\emph{pointwise rubric scoring}, not of LLM judging as such. Together with
Section~\ref{sec:protocol}, where MT-Bench's native template moves the ceiling from $0.623$
to $0.798$, the honest summary of this paper is that \textbf{protocol design dominates
panel size}, and that the generic pointwise rubric is the worst case rather than the
representative one.

\paragraph{But pairwise buys a new facet, and it is not small.} Position bias is a
systematic shift, not noise. A system placed in the first slot wins $\mathbf{8.6}$
percentage points more often than the same system, judged by the same judges on the same
items, placed second; the shift varies across systems with a standard deviation of $7.3$
points and reaches $25$ points at its worst, so it is not a constant that cancels. The
median improvement claimed in the 53 published win-rate comparisons is $7.0$ points.
\textbf{An evaluation that does not balance presentation order therefore carries a
systematic bias $1.23\times$ the median effect it is trying to detect}, which is worth
more than the entire reliability gain from adding judges.

\paragraph{The 53 deltas, no longer dropped.} With a measured pairwise floor we can price
the win-rate comparisons we previously had to set aside. Order-balanced, the MDD is $19.1$
points at our common 30 items and one judge, $10.1$ at Arena-Hard's native 500, and $9.8$
at AlpacaEval~2's native 805, falling to $7.5$ and $7.1$ at two judges. Against the floor
at native item counts, \textbf{37 of 53 ($70\%$) published win-rate improvements fall
below it} at one judge, and 27 of 53 ($51\%$) at two. The pairwise protocol is better
conditioned than the pointwise one and the claims made on it are still, in the majority,
smaller than the noise, not because the ceiling binds, but because a 7-point win-rate
difference is simply a small effect on a binary outcome measured over a few hundred items.
The caveats of Section~\ref{sec:audit} apply here too: this is our baseline, our items and
our judges, and it is a floor transferred to claims made on other instruments.

\begin{table}[t]\centering\small\setlength{\tabcolsep}{5pt}
\caption{The pairwise arm: 12 systems against a fixed \texttt{gpt-4o-mini} baseline on 24
shared items, 4 judges, 2 replicates, run in both presentation orders. The response is a
win indicator, so $\sigsq{s}$ and MDD are in win-rate units and are not comparable to the
0--5 rows elsewhere; ceiling and $\erho$ are unitless and are. ``Balanced'' is the mean of
the two orders, which is the design a careful pairwise evaluation actually runs.
``Frontier-only'' drops the 2023 models, matching the restriction of
Appendix~\ref{sec:frontier}. Under a binary preference $\sigsq{sj}$ falls two orders of
magnitude below $\sigsq{s}$ and the ceiling all but vanishes---at the cost of a position
facet that a pointwise design does not have.}
\label{tab:pairwise}
\begin{tabular}{lccccccc}
\toprule
Arm & design & $\sigsq{s}$ & $\sigsq{sj}$ & share & ceiling & $\erho$@24 & judges@0.8 \\
\midrule
System first (AB) & 11$\times$20$\times$4 & 0.0874 & 0.00034 & 33.7\% & 0.996 & 0.937 & 1 \\
Baseline first (BA) & 12$\times$19$\times$4 & 0.0705 & 0.00075 & 27.6\% & 0.989 & 0.914 & 1 \\
\textbf{Balanced} & 11$\times$19$\times$4 & 0.0801 & 0.00110 & \textbf{35.7\%} & \textbf{0.986} & 0.934 & \textbf{1} \\
\addlinespace
AB, frontier-only & 9$\times$20$\times$4 & 0.0432 & 0.00207 & 19.8\% & 0.954 & 0.858 & 1 \\
BA, frontier-only & 9$\times$19$\times$4 & 0.0442 & 0.00119 & 17.5\% & 0.974 & 0.864 & 1 \\
\midrule
\multicolumn{8}{l}{\emph{position bias:} slot-A advantage $8.6$ points, sd across systems $7.3$, max $25.0$}\\
\multicolumn{8}{l}{\emph{floor (balanced, 1 judge):} $19.1$ pts @30 items, $10.1$ @500, $9.8$ @805}\\
\multicolumn{8}{l}{\emph{53 published win-rate claims:} median $7.0$ pts; $37$ ($70\%$) below the native-count floor}\\
\bottomrule
\end{tabular}
\end{table}

\subsection{Which judges count: the primary universe}\label{sec:primary-universe}

Every reliability quantity in this paper is conditional on a universe of judges, and we
have to say which universe is the headline one. The nine-judge panel includes a
deliberately weak judge, a legacy model, and a route replica of a judge already in the
panel. Those three were put there to probe the judge facet, not because anyone would hire
them, and leaving them in the primary panel would inflate $\sigsq{sj}$ for reasons of our
own construction. A reviewer is entitled to read that as building the conclusion into the
design, so we do not ask for the benefit of the doubt: \textbf{the primary universe is the
six current-generation judges}, and the nine-judge panel is reported throughout as
sensitivity (Table~\ref{tab:universe-primary}).

The concession is real and it costs the alarm, not the argument. Restricting to the six
raises every ceiling---MT-Bench from $0.600$ to $0.798$, our screened set from $0.746$ to
$0.929$---and cuts every judge requirement, most sharply on MT-Bench, from 25 judges to 5.
Three of the four item sets now clear the conventional $0.8$ threshold. Anyone reading
this paper as ``LLM judging is hopeless'' should read the primary column instead: with a
panel of current models and a well-chosen item set, single-judge reliability above $0.9$
is achievable, and we measure it.

What does not change is the shape of the problem. The ceiling is still finite and still
set by $\sigsq{sj}$, so the item cost of a target still diverges as the target approaches
it: on MT-Bench, the strongest of the three established item sets under our rubric, a
single current-generation judge still cannot reach $0.8$ at any item count and still needs
five judges to get there at thirty items. The ordering of item sets is identical under
both universes, and the screened set remains the cell where item construction has bought
the most. Every structural claim in Section~\ref{sec:ceiling} is an algebraic consequence
of the crossed design and holds under any universe whatever; only the numbers move.

Two further points on the conditioning. First, the six are not independent draws from some
population of judges, they are two vendor families at three capability tiers, and
Section~\ref{sec:provenance} shows the family term is the larger one, so a panel of six
judges from a \emph{single} vendor is a different and worse universe than the one tabulated
here. Second, restricting the universe restricts the claim: results in the primary column
describe what a panel of current frontier judges can resolve today, and say nothing about
what the same benchmark will resolve when those judges are replaced, which on the evidence
of the legacy judge is a live concern rather than a hypothetical one.

\begin{table}[t]\centering\small\setlength{\tabcolsep}{4pt}
\caption{Primary judge universe: the six current-generation judges, with the nine-judge
panel beside it as sensitivity. The six drop the deliberately weak judge, the legacy
model and the route replica, and are the universe a practitioner assembling a panel today
would actually draw from. Quantities are as in Table~\ref{tab:main}: MDD and $\erho$ at a
common $n_i=30$, one judge, one run; ceiling is $\sigsq{s}/(\sigsq{s}+\sigsq{sj})$ with a
parametric bootstrap interval; judges is the count needed for $\erho=0.8$ at 30 items and
3 replicates. Restricting to current-generation judges raises every ceiling and lowers
every judge requirement, and three of the four item sets then clear $0.8$---but the
ordering of item sets, the finite ceiling, and the divergence of the item cost near that
ceiling are unchanged.}
\label{tab:universe-primary}
\begin{tabular}{lcccccc}
\toprule
& \multicolumn{3}{c}{primary: current-generation 6} & \multicolumn{3}{c}{sensitivity: all 9} \\
\cmidrule(lr){2-4}\cmidrule(lr){5-7}
Benchmark & ceiling & 95\% CI & judges & ceiling & 95\% CI & judges \\
\midrule
Screened (ours) & \textbf{0.929} & [0.89,\,0.95] & 1 & 0.746 & [0.66,\,0.81] & 2 \\
MT-Bench & \textbf{0.798} & [0.68,\,0.87] & 5 & 0.600 & [0.44,\,0.71] & 25 \\
Arena-Hard & \textbf{0.918} & [0.86,\,0.95] & 1 & 0.792 & [0.69,\,0.86] & 2 \\
AlpacaEval\,2 & \textbf{0.829} & [0.71,\,0.90] & 3 & 0.706 & [0.55,\,0.80] & 7 \\
\bottomrule
\end{tabular}
\end{table}

\subsection{What improvements do papers actually claim?}\label{sec:deltas}

The floor only matters relative to the effects being reported, so we measured those too.
From the audit frame (\S\ref{sec:audit}) we extracted each paper's headline improvement
on a judge-scored benchmark, together with the scale it was reported on, recovering 157
usable comparisons from 227 eligible papers. Those 157 divide into 60 reported on a
pointwise scale we can convert to a common 0--5 basis, 53 reported as win rates, and 44 on
scales we could not convert, accuracy on a bespoke subset, composite indices, or rubric
totals whose range was not stated. The 44 are excluded from every quantity below rather
than converted on a guess, which is the same discipline we apply to win rates.

Scale conversion matters and is usually left implicit. MT-Bench reports on 1--10,
several benchmarks on 0--100, ours on 0--5; a 0.2-point MT-Bench delta is 0.11 points on
a 0--5 basis. We convert pointwise scales by their range,
$\delta_{0\text{-}5}=\delta\cdot 5/(\text{hi}-\text{lo})$, and \emph{do not} convert
win rates: a pairwise win-rate difference is a different estimand, and inventing an
exchange rate would be precisely the unstated conversion this paper argues against.
They are reported separately.

Across the 60 pointwise comparisons the median reported improvement is \textbf{0.28
points} on a 0--5 basis (p25 $=0.13$, p75 $=0.63$), smaller than every floor in
Table~\ref{tab:main}. Two comparisons against those floors are available, and they differ
in how much they are worth.

\paragraph{The matched comparison.} Only one benchmark appears often enough in the
recovered deltas to be compared against its own measured floor: 17 of the 60 are
MT-Bench. Their median improvement is 0.17 points on a 0--5 basis, and \textbf{all
seventeen fall below MT-Bench's floor}, 0.63 at 30 items, and still 0.54 at MT-Bench's
native 80 items, since items saturate and the floor barely moves. Nine of the seventeen
state an item count of their own, and pricing each of those against the floor at
\emph{its own} stated count---rather than at any common one---changes nothing: the count
still below is seventeen. The saturation result is what makes this robust, since a paper
would have to have run orders of magnitude more items, not a few more, to buy itself a
materially lower floor. This is the only strictly like-for-like statement we can make, and
$n=17$ is small, but it is exact.

\paragraph{The pooled comparison, and why it is an upper bound.} Applying a single
measured floor to all 60 deltas puts 60--73\% below it, depending on which floor
(AlpacaEval~2's 0.41 at its native 805 items to Arena-Hard's 0.61 at its native 500), and
85\% below our screened set's 1.24. We report the range rather than a point, and as an
upper bound, for three reasons. Most of the 60 come from bespoke one-off instruments we
never measured, so no floor of ours is strictly theirs. Published work evaluates at native
item counts higher than our common $n_i=30$, which lowers the true floor. And
\S\ref{sec:protocol} shows our generic rubric is the \emph{conservative} case, so native
protocols would lower the floor further still. Each correction pushes the fraction down.
The defensible claim is the direction and the order of magnitude---reported improvements
sit at the scale of the measurement error, not comfortably above it---not a specific
percentage.

\section{What published work discloses}\label{sec:audit}

We audited a frame of 628 arXiv papers (2024--2026) retrieved by five preregistered
queries, fixed before any paper was read; 320 were sampled under a fixed seed, 312
coded successfully and 227 report a judged system comparison. Table~\ref{tab:audit}
reports rates over those 227. As a check on draw size, the same rates computed on an
earlier 120-paper draw and on the full 320-paper draw agree closely (replication 20.7\%
vs 19.8\%, uncertainty 45.7\% vs 45.8\%); those two percentages are over the eligible
subsets of each draw, not over the 227. Coding was performed by a pinned model at temperature
zero, constrained to quote verbatim evidence or answer \texttt{not\_stated}, with
inference forbidden; hand validation of spot-checked fields agreed at $\approx$88\%.
Two earlier regex-based coders were discarded after validation showed they could not
distinguish training runs from evaluation runs, or a data-source model from a judge.

\begin{table}[t]\centering\small
\caption{Disclosure among 92 papers that report a judged system comparison.}
\label{tab:audit}
\begin{tabular}{lr}
\toprule
Disclosure & Rate \\
\midrule
Names the judge model & 91.2\% \\
States the number of evaluation items & 70.0\% \\
Reports any uncertainty (CI / SD / test) & 45.8\% \\
Pins a judge version & 38.8\% \\
Controls presentation order & 15.9\% \\
States whether the evaluation was repeated & \textbf{19.8\%} \\
\bottomrule
\end{tabular}
\end{table}

\paragraph{Do the checkable claims survive?} Disclosure being absent is not the same as
conclusions being wrong, so we close the loop where the data allows. Of the 60 pointwise
improvements recovered, 17 are on MT-Bench, the one benchmark whose floor we measure
directly. Setting each against MT-Bench's floor at its own native 80 items,
\textbf{none of the seventeen exceeds it}. Thirty-nine of the 60 papers also state their
item count, so the same test could be extended once floors exist for their benchmarks; we
report the matched subset rather than transferring a floor across instruments.

Combined with \S\ref{sec:floors} and \S\ref{sec:deltas}, the field is reporting
differences well below its instruments' resolution while omitting the information needed
to notice. The coding schema, both automated codings, the human coding and the codebook
are available from the author on request.

\section{Recommendations}

\begin{enumerate}\itemsep2pt
\item \textbf{Report the judge universe.} A reliability or significance claim is
undefined without it.
\item \textbf{Use the benchmark's native protocol, and say which you used.} This is the
cheapest lever we found: it costs nothing per call, and on matched material it moved
MT-Bench's ceiling from 0.623 to 0.798 and the judges needed for $\erho=0.8$ from 18 to 3.
Rubric, scale and reference answers are design variables, not incidental formatting.
\item \textbf{Spend on judges before items.} Past a modest item count the marginal
return on items is near zero; judges are the binding axis.
\item \textbf{Report MDD alongside the delta.} It is precisely estimable where
reliability coefficients are not.
\item \textbf{Screen items for discrimination} when constructing benchmarks.
\item \textbf{Disclose replication, and whether the inference path caches responses.}
\end{enumerate}

\section{Limitations}\label{sec:limits}

The system panels behind Table~\ref{tab:main} differ between rows in two ways. The screened
set omits \texttt{claude-opus-5} and \texttt{claude-sonnet-5}, unavailable as systems at
collection time, so its base panel is 36 rather than 42; and it carries 23 null-arm
replicas the public sets do not, giving 59. Refitting without the replicas moves nothing
material (ceiling 0.735 against 0.746, MDD 1.25 against 1.24), but the rows are matched on
judges, rubric and scale rather than on systems, and cross-set statements should be read
with that in mind.

\paragraph{Judge panels.} The closed panel spans two model families; a third was
unavailable throughout, so panel-composition generality is untested beyond the splits in
\S\ref{sec:universe} and \S\ref{sec:provenance}. The open-weight panel is five models at a
single quantisation (q4\_K\_M for the larger three) served locally at a 4096-token context.
Quantisation and serving configuration are plausible facets we did not vary, so the
provenance contrast is between \emph{these} panels rather than between open and closed
weights in general. Qwen3~8B was excluded after 13.0\% of its calls returned empty
completions in thinking mode, leaving one family represented by a single model.

\paragraph{Protocol.} Items are scored under a common rubric rather than each benchmark's
native protocol. That isolates the item effect but states results for a standard pointwise
protocol applied to those items. We replicate MT-Bench natively (\S\ref{sec:protocol}) and
find the generic rubric the conservative case, and we now measure a pairwise protocol
directly (\S\ref{sec:pairwise}), which is the form Arena-Hard and AlpacaEval~2 actually
take. That arm is not those benchmarks as deployed: it uses our items, our judges and a
single fixed baseline, rather than Arena-Hard's own prompt set, template and reference
model. So the gap for those two rows is now bounded rather than closed, we can say what a
pairwise protocol does to the ceiling in general, but not what each benchmark's specific
pipeline does to it. Given that the pairwise arm moves the ceiling more than any other
manipulation in this paper, this remains the largest source of uncertainty about how our
numbers transfer, and the most useful thing a replication could pin down.

\paragraph{Data quality.} The Arena-Hard cell rests on 17 items against 32--43 for the other sets.
This is a collection shortfall rather than an attrition one: the harvest recorded no errors and no
dropped cells, and the source pool holds 500 prompts, so the constraint was budget rather than the
benchmark. Its $\sigsq{s}$ is accordingly the least precisely estimated of the four, and its ceiling
carries the widest interval in Table~\ref{tab:main}. Approximately 15\% of
generations were truncated at the token cap, which penalises verbose systems consistently
across judges.

\paragraph{Scope of the human and audit comparisons.} The SummEval comparison is one
dimension of one dated summarisation task and plausibly flatters judges. The audit's
replication field conflates repetition for variance with repeated calls for order control,
so 19.8\% is a conservative upper bound.

\section{Conclusion}

Judged benchmarks have a resolution and it is measurable. Because the system-by-judge
interaction carries no dependence on item count, a single-judge protocol has a ceiling
that more data cannot raise. That much is algebra and holds for any judge panel, any
rubric, any benchmark.

Where that ceiling sits is not algebra, and we found it to be strongly conditional. Under
a nine-judge universe that includes a legacy and a deliberately weak judge, none of our
four item sets reaches conventional reliability at any item count; under six
current-generation judges, three of the four do. Switching MT-Bench from a generic rubric
to its own native protocol moves its ceiling from 0.623 to 0.798 on matched material. A reliability number for
a judged benchmark is therefore not a property of the benchmark. It is a property of the
benchmark, the judge universe and the protocol together, and reporting it without the
latter two leaves it undetermined rather than merely imprecise, which is what nearly all
published work currently does.

Two things follow that do not depend on the conditions. The exchange rate between judges
and items is steep enough to reverse the usual instinct, and it steepens as the target
rises: two judges at thirty items are worth 77 items at one judge at $\erho=0.75$, and
past a modest count items are nearly worthless, the count needed diverges as the target
nears whatever ceiling the protocol produces. And the improvements the field reports sit at the scale of the
measurement error rather than comfortably above it, every one of the seventeen MT-Bench
improvements we recovered falls below MT-Bench's own floor. The limit is not a failure of
LLM judges, which are more self-consistent than human experts on a matched design, but a
property of the measurement design, which binds human raters equally.

\appendix

\section{Supporting analyses}\label{app:supporting}

Each result below is stated with the numbers that carry it. Full per-cell tables are in the
released artifacts (Appendix~\ref{app:artifacts}).

\paragraph{The judge universe.}\label{sec:universe} $\erho$ is defined relative to a universe of admissible
judges, and published work never states one. Recomputing the ceiling under four choices
moves MT-Bench from $0.600$ (all nine) to $0.649$ (drop legacy), $0.691$ (drop weak) and
$0.798$ (current-generation six); our screened set moves $0.746\to0.804\to0.840\to0.929$,
Arena-Hard $0.792\to0.918$, AlpacaEval~2 $0.706\to0.829$. The ordering is stable, the level
is not, so a reliability figure reported without its judge universe is uninterpretable.
Section~\ref{sec:primary-universe} explains which one we take as primary.

\paragraph{Route versus model identity.} The route replica serves the same weights as
\texttt{anth-cheap} through a different provider endpoint, so the variance of their
interaction difference isolates route. It is $0.0010$--$0.0074$ across the four item sets,
or $1.4\%$--$5.0\%$ of the disagreement distinct current-generation models produce on the
same systems. $\sigsq{sj}$ is about \emph{which model judges}, not where it is served; a
panel of one model behind several endpoints buys almost none of the independence a panel of
distinct models buys.

\paragraph{Judge provenance is a facet.}\label{sec:provenance} Five open-weight judges on identical material
disagree $2.17\times$ more across model families than within them once scale usage is
standardised away ($2.92\times$ on raw scores, before standardisation, the difference is a
scale-compression artefact and we report the standardised figure). Mixing families buys
independence but not a smaller panel.

\paragraph{Discriminating power at the frontier.}\label{sec:frontier} Published comparisons are between closely
matched frontier systems, not across a 2024--2026 capability range. Restricted to ten
current frontier models, system variance falls to $1.9\%$ on our screened items and $2.8\%$
on MT-Bench. Arena-Hard is the exception at $15.7\%$, consistent with its design intent and
the one benchmark still separating frontier systems.

\paragraph{Is the screened set circular?} Our items were selected for between-system
variance, so finding it afterwards would be guaranteed. Two checks. The screener
(\texttt{gpt-4.1}) is deliberately outside the evaluation panel, and splitting that panel by
vendor family shows the screened set lifts the system share $3.16\times$ over MT-Bench for
OpenAI-family judges and $3.82\times$ for Anthropic-family, the gain is \emph{larger} on the
family least like the screener, the opposite of selection on one model's taste. And ranking
items by discrimination within random halves of the system panel and correlating the two
rankings gives a split-half Spearman $\rho=0.757$ (95\% range $0.413$--$0.893$), the highest
of the four item sets, against $0.597$ for MT-Bench and $0.551$ for Arena-Hard. Screening
captured a transferable property, not the screener's preferences.

\paragraph{What counts as a system.} The 42 systems are 14 models $\times$ 3 prompt
configurations, so configurations are nested in models rather than crossed. Fitting the
nested model, configuration-given-model carries $55.7\%$ of system variance on MT-Bench,
$46.7\%$ on AlpacaEval~2, $26.1\%$ on Arena-Hard and $0\%$ on our screened items. Treating
the model as the object of measurement lowers MT-Bench's ceiling from $0.600$ to $0.404$ and
AlpacaEval~2's from $0.706$ to $0.567$. Whether a benchmark can separate \emph{models} is a
harder question than whether it can separate \emph{system configurations}, and the results in
the body answer the easier one.

\paragraph{A panel and an ensemble are the same object.} Almost nobody reports $n_j$ separate
scores; they average the panel into one. G-theory says the mean of $k$ judges carries
$\sigsq{sj}/k$, which is falsifiable. A single ensemble judge cannot be fitted---at $n_j=1$
the judge facet is unidentified---so we form \emph{disjoint} ensembles of size $k$, refit on
those $\lfloor n_j/k\rfloor$ ensemble judges, and compare the resulting $\sigsq{sj}$ against
the prediction. Over 60 random partitions per cell at $k=2$ and $k=3$ on all four item sets,
observed over predicted ranges $0.93$--$1.11$ with median $1.01$. Every judge count in this
paper may be read as an ensemble size.

\paragraph{Temperature is not separated.} All judges run at a fixed decoding temperature, so
what the replicate facet measures is run-to-run variation \emph{at} that temperature, which
is $\sigsq{e}$ and not a temperature effect. Separating it needs an arm crossing temperature
with judge, which we have not run. This is the one place where ``judge'' remains a bundle.

\paragraph{Reliability is not validity.} $\erho$ asks whether a panel reproduces itself, not
whether it is right. On SummEval \citep{fabbri2021summeval}, moving from $(n_i{=}10,n_j{=}1)$ to $(n_i{=}50,n_j{=}9)$
raises $\erho$ from $0.711$ to $0.959$ ($+0.248$) while Kendall $\tau$ against the
three-expert consensus rises only from $0.596$ to $0.710$ ($+0.114$) and is flat over the
last four configurations. Recovery of the human top three sits at $23\%$--$32\%$ across every
configuration against a $18.8\%$ chance baseline, and does not improve on either axis.
Reliability is a precondition for a believable comparison, not a substitute for one.

\paragraph{The limit is structural, not a judge deficiency.}\label{sec:human} On SummEval's fully crossed
human design (16 systems $\times$ 100 articles, 3 experts and 5 crowd workers), rater-by-system
variance per unit of system signal is $0.07$--$0.13$ for LLM judges against $0.34$ for experts
and $0.67$ for crowd workers. Two judges match three experts; crowd workers produce
essentially no system-level signal ($\sigsq{s}\approx0.002$ at five raters). Substituting
humans makes the resolution worse at far greater cost.

\paragraph{What does not explain the floor.}\label{sec:confounds} Type~I error is correctly calibrated: on pairs
with a true difference of exactly zero---two independent generations of the same system---the
modal protocol's false-positive rate is $2.5\%$, below nominal. The problem is power, not
calibration. Self-preference is real but small, explaining $0.8\%$--$4.8\%$ of $\sigsq{sj}$.
Length explains more: within-item correlation between response length and score is
$0.21$--$0.25$ on the public benchmarks and length explains $13\%$--$17\%$ of between-system
variance, so part of what those benchmarks measure as quality is verbosity. On our screened
items, which carry explicit format and length constraints, the correlation is $-0.004$, item
design can remove this.

\paragraph{What the floor costs in practice.} Three consequences a practitioner can act on.
\emph{The naive bootstrap understates the interval.} Resampling items only---what a careful
author does today---gives a half-width of $0.34$--$0.49$ points against a true single-judge
MDD of $0.58$--$1.24$, understating by $1.3\times$ to $2.8\times$ because it ignores the judge
facet entirely. \emph{Rank inversions are near chance for adjacent systems.} For two systems
adjacent in the true top ten, whose mean separation is $0.010$--$0.013$ points, the
probability of returning them in the wrong order is $45\%$--$49\%$ at one judge and barely
moves at three. \emph{Under a call budget the optimum is interior.} Minimising $\sigsq{\delta}$
subject to cost $\propto n_i n_j n_r$ gives, on MT-Bench, $30\times4\times1$ at 120 calls,
$60\times6\times1$ at 360 and $108\times10\times1$ at 1080; on our screened items
$20\times6\times1$, $36\times10\times1$ and $60\times18\times1$. The optimal judge count grows
with budget rather than saturating, and a single run is optimal at every budget examined.

\section{Reading the main tables}\label{app:reading}

\paragraph{Raw points do not rank item sets.} The four sets use different amounts of the
scale, so a raw MDD is measured against a different ruler in each. System means span $2.63$
points on our screened set against $1.09$ on MT-Bench and $1.18$ on AlpacaEval~2. In units
of each set's own between-system spread the ordering changes: $1.57$~SD for Arena-Hard,
$1.77$ for the screened set, $2.31$ for AlpacaEval~2, $2.67$ for MT-Bench, so the screened
set has the largest raw MDD and the second smallest standardised one. This is the same
scale-compression artefact we document for judge panels in \S\ref{sec:provenance}: any
absolute-scale comparison of noise rewards an instrument for compressing its output. Raw
points are the right unit for comparing a benchmark's floor against improvements reported
\emph{on that benchmark} (\S\ref{sec:deltas}), and the wrong unit for ranking benchmarks
against each other.

\paragraph{What screening bought, and what it cost.} Screening raised $\sigsq{s}$ by an
order of magnitude over MT-Bench ($0.456$ against $0.044$) and roughly doubled the usable
scale range, but raised the interaction terms alongside it ($\sigsq{si}$ $0.474$ against
$0.286$; $\sigsq{sij}$ $0.683$ against $0.267$). That is unsurprising on reflection: items
selected for separating systems will also separate them in \emph{different orders}, which is
system-by-item interaction by definition. The net is a single-judge ceiling of $0.746$ and a
standardised MDD of $1.77$~SD, in both cases second to Arena-Hard's $0.792$ and $1.57$.
Screening buys a better instrument than the two weakest public sets and does not beat the
strongest: a lever worth pulling, not a solved problem.

\paragraph{Three MDD figures, three questions.} MDD@30 ($0.58$--$1.24$) evaluates every set
at a common thirty items, which is what makes them comparable. MDD at native counts
($0.41$--$1.24$) evaluates each at the count its users deploy, the right figure against a
\emph{published} result. MDD at infinite items ($0.40$--$1.09$) is the floor no item budget
can beat under one judge. All three are single-judge, single-run, and none moves much,
because items saturate.

\paragraph{The object of measurement.} $\sigsq{s}$ is defined over a \emph{population of
systems}, exactly as $\erho$ is defined over a universe of judges. ``MT-Bench has 4.5\%
system variance'' is shorthand for ``across the 42 systems studied here''. We report the
near-frontier subset separately (\S\ref{sec:frontier}) precisely because the choice of
system population changes the answer: discriminating power is a property of the pair, not of
the benchmark alone. Neither facet should be reported without its universe.

\section{Audit and infrastructure detail}\label{app:auditdetail}

\paragraph{Validation.} Coding reliability was established in two stages. The sample was
first coded a second time by a different model family (GPT-5.6) under an independently
worded prompt. A 30-paper subsample was then coded by a human working \emph{blind}---no
automated answers, no evidence quotes---so that agreement measures coding difficulty
rather than suggestibility.

Against the human, overall agreement is 88.0\% ($\kappa=0.73$). Crucially, the field the
headline depends on validates almost perfectly: whether the evaluation was repeated
reaches $\kappa=0.89$ with a single disagreement in thirty papers, the human coding
16.7\% and the automated coder 20.0\%. Whether the scorer was named is perfect
($\kappa=1.00$). Uncertainty reporting is substantial ($\kappa=0.65$).

The human coding also resolves an ambiguity the two automated coders could not. Between
themselves they agreed only moderately on replication ($\kappa=0.47$, estimating 15\% and
29\%), which we had provisionally reported as a range. The human agrees closely with the
first coder and not the second, indicating that the disagreement was one coder in error
rather than genuine ambiguity in the papers. We therefore report the lower, validated
figure.

One field remains genuinely ambiguous and is reported as a range: version pinning
($\kappa=0.45$; 53\% automated, 73\% human), where the human applied a more generous
standard for what counts as a pinned version. Item counts show 86.7\% agreement, though
$\kappa$ is undefined there because the human coded every paper affirmatively.

\paragraph{Closing the loop: does disclosure predict survival?} The audit so far shows that
disclosure is absent, not that any conclusion was wrong, and those are different
complaints. The link we can actually test is whether the papers that \emph{do} report
enough to be checked are also the ones whose claims clear a floor. Of the 227 papers
reporting a judged comparison, $146$ ($64.3\%$) state both a judge model and an item count,
which is the minimum needed to compute an implied floor at all; the remaining third cannot
be checked by anyone, including their own authors.

Among the 60 recovered deltas, 35 come from papers stating both and 25 do not. Applying the
same MT-Bench floor to both groups at each paper's own stated item count, $40\%$ of the
fully-disclosing claims exceed it against $24\%$ of the rest. The absolute level here is not
trustworthy---it transfers one instrument's floor to claims made on others, which is the
error this paper argues against---but the \emph{contrast} is, because both groups are priced
against the identical floor, so whatever bias the transfer introduces is common to them. The
direction is worth stating plainly: better-disclosed claims are not the fragile ones.
Disclosure is not what makes a result survive, but it does travel with results that do,
most likely because papers that report an item count tend to report a larger one.

This is the causal link between Section~\ref{sec:audit} and Section~\ref{sec:floors}, and it
cuts against the most cynical reading of the audit. The problem is not that the field is
making claims it knows to be unsupported; it is that two-thirds of a corpus can be checked
and one-third cannot, and no reader can tell from the outside which third a given paper is
in. That is a reporting failure with a cheap fix, and it is the one recommendation in this
paper that costs nothing to adopt.

\section{Estimation notes}\label{app:estimation}

\paragraph{Where the two-system difference comes from.} Two systems measured on the
\emph{same} items and judges share those facets, so the item and judge main effects cancel
in the difference and only the interactions with system survive. That is why
$\sigsq{\delta}$ carries $\sigsq{si}$, $\sigsq{sj}$, $\sigsq{sij}$ and the residual but not
$\sigsq{i}$, $\sigsq{j}$ or $\sigsq{ij}$: this is relative error, appropriate for ranking or
comparing systems within a study, and not absolute error, which is what one would need to
compare a score against a fixed external standard. The variance of the difference is
$2\sigsq{\delta}$ because the two systems are independent draws from the system population
given the shared design.

\paragraph{The 1.96 factor.} $\mathrm{MDD}_{95}\approx1.96\sqrt{2\sigsq{\delta}}$ assumes
the difference of two system means is approximately normal. With $n_i\ge20$ items the
central limit theorem is doing the work and the approximation is unremarkable; at very
small $n_i$ it will be optimistic, and a $t$ quantile on the appropriate error degrees of
freedom should be substituted.

\paragraph{ANOVA rather than REML, and negative estimates.} We estimate components by
expected mean squares on the complete crossed block. The estimator is unbiased and closed
form, which matters here because we refit under many judge universes, system populations
and item subsets, and a closed form makes those refits cheap and deterministic. Its known
cost is that individual component estimates can go negative when the true value is near
zero; we truncate at zero when reporting variance \emph{shares} and leave the raw value in
place when computing $\sigsq{\delta}$, so that truncation never flatters a floor. REML
would avoid negative estimates and is the better choice for a single definitive fit, but it
is iterative and would make the bootstrap over 600--800 resamples materially more
expensive. Where we report intervals we use a parametric bootstrap rather than Wald
intervals, precisely because the sampling distribution of a variance component near the
boundary is not symmetric.

\paragraph{Degrees of freedom on $\sigsq{s}$.} The system facet has $n_s-1$ degrees of
freedom, so with 42 systems $\sigsq{s}$ is the least precisely estimated component in the
design, and it is the numerator of every reliability quantity. This is the main reason we
report ceilings with bootstrap intervals rather than as point estimates, and why
Arena-Hard---which carries the fewest items and therefore the widest interval---is the cell
we are least willing to make claims about.

\section{Additional figures}\label{app:figures}

\begin{figure}[H]\centering
\includegraphics[width=.94\linewidth]{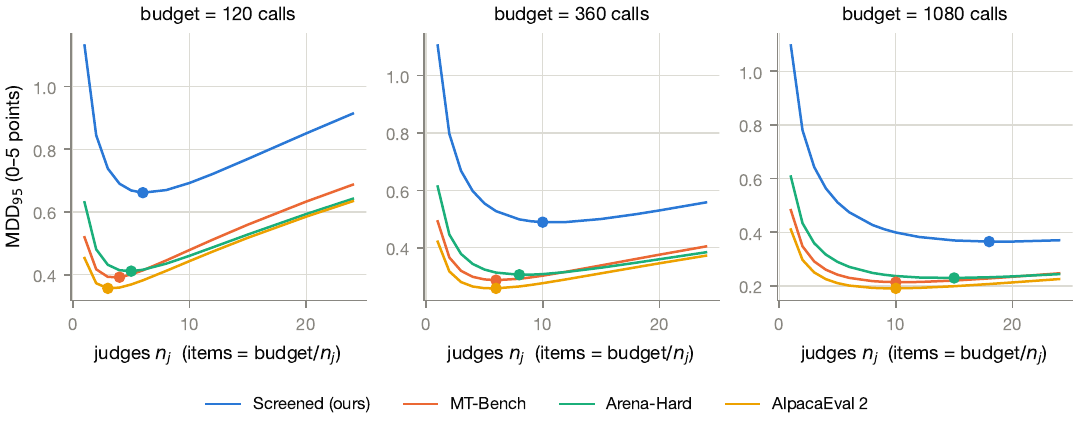}
\caption{Iso-cost allocation. At a fixed budget of $n_i n_j n_r$ calls, MDD against how the
budget is split; the marked point is the optimum. It is interior at every budget, and it
moves toward more judges as the budget grows rather than saturating.}
\label{fig:isocost}
\end{figure}

\begin{figure}[H]\centering
\includegraphics[width=.90\linewidth]{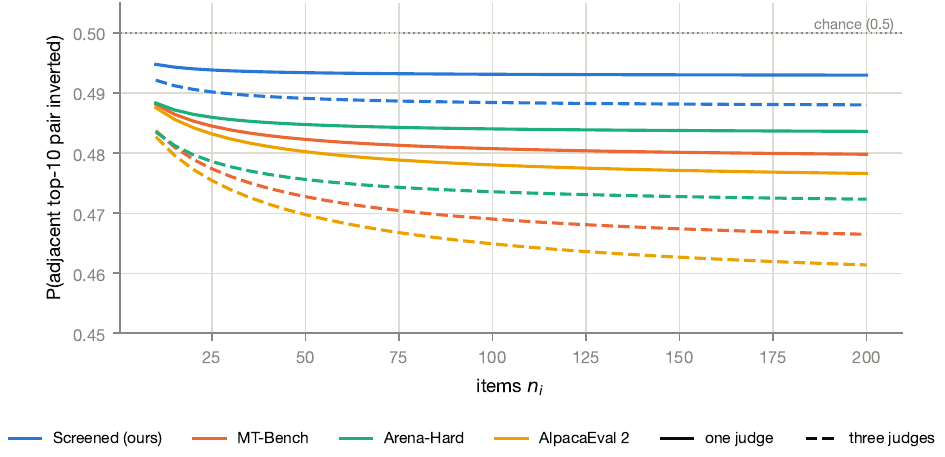}
\caption{Probability that two adjacent systems in the true top ten are returned in the
wrong order, against item count, at one and three judges. The curves flatten well short of
the 0.5 chance line: items buy rank stability slowly, and a third judge buys more than
doubling the items.}
\label{fig:rankstab}
\end{figure}

\begin{figure}[H]\centering
\includegraphics[width=.94\linewidth]{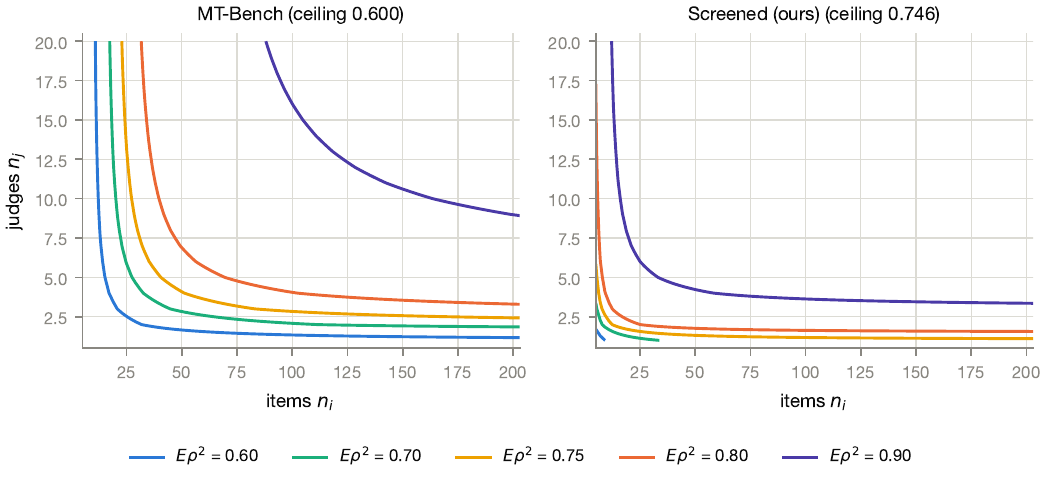}
\caption{Sizing nomogram. Contours of $\erho$ over judges and items; read off the panel a
target requires. Each contour flattens to a horizontal asymptote, and below that judge
count no item budget reaches the target, the ceiling, drawn.}
\label{fig:nomogram}
\end{figure}

\section{Artifacts}\label{app:artifacts}

\paragraph{Sieve: the discrimination-screened item set.} The screened row throughout this
paper is a releasable benchmark, named here so it can be cited and criticised separately
from our results. \textbf{Sieve} is 38 open-ended instruction items selected for
between-system discriminating power, constructed as follows. \emph{Pool:} 71 items across
four strata designed to force systems apart rather than to sample a task
distribution, \emph{pushback} (24 items, a false premise or unsafe-but-plausible ask),
\emph{compose} (18, interacting constraints), \emph{constraint} (16, a checkable format or
length rule), \emph{verifiable} (13, a checkable factual or arithmetic core).
\emph{Screening design:} $5$ systems $\times$ 71 items $\times$ 1 judge $\times$ 2
replicates, 720 calls, on the same holistic 0--5 rubric used in the main study; a checklist
variant discriminated roughly $4\times$ worse and was dropped. \emph{Screening judge:}
\texttt{openai/gpt-4.1-2025-04-14}, deliberately \emph{not} a member of the nine-judge
evaluation panel, so no item is selected using a judge that later scores it.
\emph{Selection rule:} retain items with between-system variance $\ge1.0$ on the 0--5 scale;
pool mean was $3.99$ (sd $1.36$) and 38 of 71 cleared it. The main study harvests 30 of the
38. Sieve is a measurement instrument, not a capability benchmark: items were chosen because
systems answer them differently, not because they represent any task distribution.

\paragraph{Artifacts.} The crossed judgment dataset (373{,}019 calls across four item sets,
nine judges, three replicates, with per-call model, prompt configuration, replicate index,
raw completion and parsed score); Sieve, its stratum labels and per-item discrimination
scores, plus the 33 screened-out items; the pairwise arm (4{,}608 preference judgments in
both orders); a D-study calculator that takes variance components and returns item and judge
requirements; and the audit, 628-paper frame, both automated codings, the blind human
coding, and the codebook.

\paragraph{Licensing and reproduction.} Data and item sets under CC~BY~4.0, code under MIT.
Judgment records carry the dated model identifier requested rather than the alias, since the
proxy echoes the alias back; readers reproducing against another endpoint should expect
different snapshots behind the same names. The endpoint used here enforces a client
allowlist at the edge, so an external reader cannot replay the harvest against it and must
supply their own provider credentials.

\paragraph{Availability.} The datasets and code described above are available from the
author on request.

{\small
\bibliographystyle{plainnat}
\bibliography{refs}
}

\end{document}